\pdfoutput=1
\documentclass[11pt]{article}

\usepackage{acl}

\usepackage{times}
\usepackage{latexsym}
\usepackage{graphicx}
\usepackage{booktabs}
\usepackage{multirow}
\usepackage{enumitem}
\usepackage{caption}
\usepackage{subcaption}

\usepackage[T1]{fontenc}
\usepackage[utf8]{inputenc}

\usepackage{microtype}

\usepackage{inconsolata}

\title{A Morphology-Based Investigation of Positional Encodings}

\author{Poulami Ghosh$^{\ast}$, Shikhar Vashishth$^{\dagger}$, Raj Dabre$^{\ddagger}$, Pushpak Bhattacharyya$^{\ast}$\\ { }
$^{\ast}$IIT Bombay, India, $^{\dagger}$Google Research, India $^{\ddagger}$NICT, Japan\\
\texttt{\{poulami, pb}\}@cse.iitb.ac.in}

\begin{document}
\maketitle
\begin{abstract}
Contemporary deep learning models effectively handle languages with diverse morphology despite not being directly integrated into them. Morphology and word order are closely linked, with the latter incorporated into transformer-based models through positional encodings. This prompts a fundamental inquiry: \textit{\textit{Is there a correlation between the morphological complexity of a language and the utilization of positional encoding in pre-trained language models?}} In pursuit of an answer, we present the first study addressing this question, encompassing 22 languages and 5 downstream tasks. Our findings reveal that the importance of positional encoding diminishes with increasing morphological complexity in languages. Our study motivates the need for a deeper understanding of positional encoding, augmenting them to better reflect the different languages under consideration.
\end{abstract}

\section{Introduction}
Pre-trained language models (PLMs) \cite{devlin2018bert,liu2019roberta,radford2019language,raffel2020exploring,brown2020language} built upon transformers~\cite{vaswani2017attention} have achieved ground-breaking results across a wide spectrum of language processing tasks such as natural language inference \cite{best_nli}, text classification \cite{class_best}, named entity recognition \cite{best_nli}, and part-of-speech tagging \cite{pos_best}. However, only a few models take into account various linguistic aspects and theories in their design  \cite{nzeyimana2022kinyabert,park2021morphology}. Morphology and word order of a language are closely related \cite{sapir1921language,comrie1989language,blake2001case}; the latter is incorporated into transformer-based models through positional encoding (PE) \cite{dufter2022position}. As language models are being developed for more languages which significantly differ in morphological typology, it is could be beneficial to construct language models that are sensitive to these linguistic nuances.  Moreover, the enormous computational cost incurred during their training is a major challenge in the development of PLMs. Acquiring a deeper understanding of how various components of a PLM function in different languages can provide valuable insights regarding their necessity across languages. This motivates us to investigate the relation between positional encoding and morphology, which is essential for wider usage of PLMs across different languages. Our contributions are:\\
\noindent\textbf{1.} We perform the first study about the varying importance of positional encoding across languages with different morphological complexity.\\
\noindent\textbf{2.} We show that the impact of PE diminishes as the morphological complexity of a language increases.\\
\noindent\textbf{3.} We conduct exhaustive experiments covering 22 different languages across 9 language families and 5 diverse natural language processing tasks.

\begin{figure*}[t]
	\centering
	\includegraphics[width=0.93\textwidth]{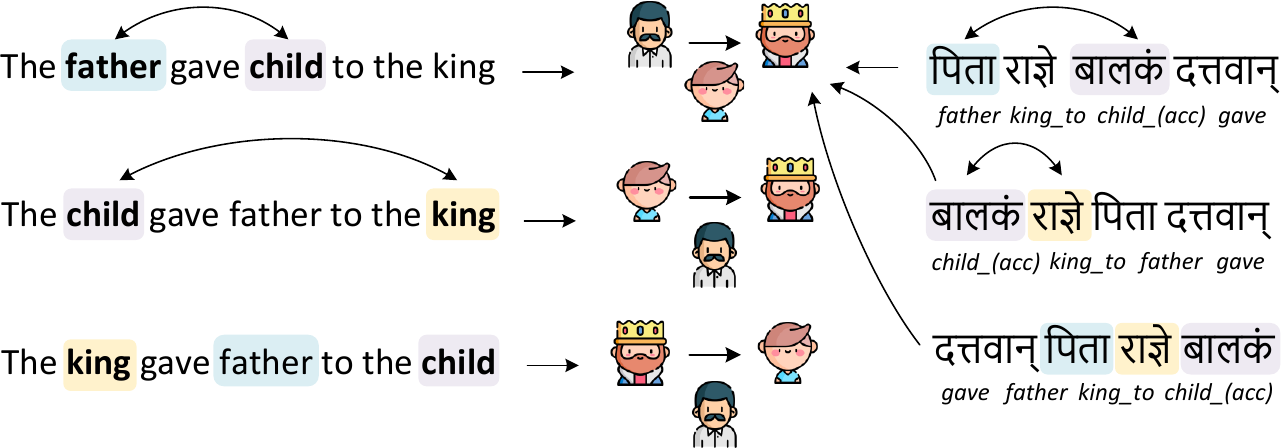}
% 	\vspace{-0.2in} 
	\caption{\label{fig:overview} The figure illustrates the effect of word order on semantics for two languages: English (\textbf{left}) and Sanskrit (\textbf{right}). English is a morphologically poor language with SVO word order whereas Sanskrit is a morphologically rich language with no dominant word order (NODOM). Distorting the word order completely alters the meaning for English. However, for Sanskrit the meaning remains intact.}
% 	\vspace{-0.15in}
\end{figure*}

%The vast diversity of morphological systems exhibited in languages worldwide constitutes an intriguing dimension of human language.\noindent 

%The structure of the paper is outlined as follows.  In section \ref{sec:related_work}, we provide a detailed overview of pre-trained language models and positional encodings, exploring their connection to linguistics. We describe the correlation between morphology and word order in section \ref{sec:morph_word_order}. Our experimental setup, including details of tasks, datasets, and language models, are outlined in section \ref{sec:experimental_setup}. Finally, in section \ref{sec:results}, we discuss the results based on the study of morphological typology.

\section{Related work}\label{sec:related_work}
%In this section, we provide a comprehensive overview of pre-trained language models, positional encodings and how they relate to linguistics.\\
\textbf{Positional Encoding (PE):} Various methods have been proposed to incorporate position information in transformer models. Absolute positions in a sequence, represented by fixed \cite{vaswani2017attention} or trainable encodings \cite{gehring2017convolutional,devlin2018bert,radford2019language,alberta}, are typically added to input embeddings. Relative positions are encoded by directly adding position biases into the attention matrix \cite{shaw2018self,yang2019xlnet,raffel2020exploring,huang2020improve,he2020deberta,press2021train}. \citet{su2021roformer} introduce rotary positional embeddings, employing a rotation matrix to encode both absolute and relative position information. Our study focuses on BERT models, which use learnable absolute PEs \cite{wang2020position,huang2020improve}. We draw insights from linguistics theories and question the design choices for  BERT-style models that were designed with English in mind.\\ %While these studies primarily focus on enhancing positional embeddings, recent papers have also studied the effect of removing positional encoding in PLMs.\\ %While these studies primarily focus on enhancing positional embeddings, investigating their removal can provide valuable insights into model behavior and potential enhancements.\\
\textbf{Absence of Positional Information:} Eliminating positional encoding results in a bag-of-words representation. \citet{sinha-etal-2021-masked} pre-train a RoBERTa model without positional embeddings and observe degraded performance on GLUE and PAWS tasks. \citet{haviv-etal-2022-transformer}; \citet{le-scao-etal-2022-language} demonstrate that causal language models lacking explicit PE remain competitive with standard position-aware models. Additionally, \citet{haviv-etal-2022-transformer} find that a pre-trained RoBERTa large model without PE exhibits higher perplexities than position-informed models. However, all these observations are limited to the English language.\\
%\textbf{Positional Embeddings in BERT:}
%Our study focuses on BERT models, which utilize learnable absolute position embeddings. \citet{wang2020position,huang2020improve} systematically investigate the effects of various positional encoding in BERT models across GLUE and SQuAD tasks. \cite{wang2020position} propose three properties of PEs, namely translation invariance, monotonicity, and symmetry, to probe BERT trained with different variants of PE and their combination.\\
\textbf{Linguistic Information in BERT:} Several works studied the linguistic knowledge encoded in PLMs such as BERT, focusing on different aspects of linguistics such as syntax \citep{goldberg2019assessing,jawahar-etal-2019-bert}, semantics \citep{ethayarajh-2019-contextual} and morphology \citep{edmiston2020systematic}. \citet{tenney-etal-2019-bert,puccetti-etal-2021-bert} investigate the extent and organization of the linguistic information encoded in BERT. \citet{gerz-etal-2018-relation} investigate the connection between language modeling and linguistic typology across 50 different languages. However, they do not consider PLMs. \citet{otmakhova-etal-2022-cross} examine how various layers within a BERT model encode morphology and syntax.%, namely English, Korean, and Russian %In light of these studies, our work explores the relationship between PE and morphology.\\

\section{Methodology}
In our work, we first quantify morphological complexity, and then systematically study the effect of removal of positional encodings during fine-tuning. Please refer to Appendix \ref{sec:morph_word_order} for details on linguistic theories governing our study.
\subsection{Quantifying Morphological Complexity}\label{subsec:quantifying_morphological_complexity}
Following \citet{kettunen2014can,jayanthi-pratapa-2021-study,ccoltekin2023complexity}, we employ type-token ratio (TTR) as an empirical proxy of morphological complexity of a language. We use the many-to-many multilingual Flores-200 benchmark \cite{costa2022no} to ensure information consistency across languages. As Chinese is an unsegmented language,  we use character level ELMo model from the pywordseg library \cite{chuang2019robust} to split Chinese text into words. The remaining languages are space-delimited. Please refer to Appendix \ref{sec:ttr_morph} for more details. %We provide more details in Appendix \ref{sec:morphological_complexity}.

%For space-delimited languages and approaches for unsegmented languages
%For word segmentation in Chinese,\footnote{Unlike languages like English and French were words are space-delimited Chinese is an unsegmented language.}
%\footnote{\url{https://pypi.org/project/pywordseg/}}.

\subsection{Morphology-based Investigation}\label{subsec:morphology_based_investigation}

To study the impact of positional embeddings, we set them to 0, effectively nullifying its effect during fine-tuning. We posit that for morphologically rich languages like Sanskrit, this would have minimal impact on downstream performance. For example, as depicted in Figure \ref{fig:overview}, the semantic meaning of a sentence in Sanskrit remains consistent even when the order of tokens is shuffled. However, this does not hold for morphologically poor languages. 

%\begin{itemize}[itemsep=2pt,parsep=2pt,partopsep=2pt,leftmargin=*,topsep=2pt]
%\item \textbf{Nullifying Positional Encoding:} 
%In this strategy, we set positional embedding in the language model to null, i.e., $\bm{p}_i = \bm{0}$. Thus, making model invariant to the token word of the input sequence. We hypothesize that for morphological rich languages such as Sanskrit this would not have much impact on the downstream performance. For instance, in Figure \ref{fig:overview}, we demonstrate that for Sanskrit, the meaning of the sentence remains consistent even when the token order is shuffled. However, this does not hold for morphologically poor languages such as English. 
% In the first setting, we preserve the positional information $p_i$ and run all analysis on a position-aware model. In the second setting, positional encoding is nullified by setting $p_i=0$ for all positions $i$.

%\item \textbf{Permuting Word Order: }
%Here, we distort the word order by shuffling a sentence at the level of uni-gram, bi-gram and tri-gram. The primary objective is to introduce varying degrees of distortion and examine the resulting loss in information across morphologically diverse languages. The proposed permutation strategy can be applied to study the robustness of the language models to the token order of the input.
%\end{itemize}

\section{Experimental Setup}\label{sec:experimental_setup}
%\subsection{Selection Criteria}\label{subsec:selection_crtiteria}
To ensure the generalizability of our findings, we choose to perform a comprehensive study spanning different languages and tasks. %In this section, we outline the principles for selecting diverse tasks and languages and provide details on the datasets and models used in evaluation.

\subsection{Tasks and Languages}
%The main objective of our research is to analyze the effect of positional encoding in languages with varying morphological complexity.
%As discussed in Section \ref{sec:related_work}, positional encoding encodes word order and thereby the syntactic relationships inherent in a language. 
As our work deals with the interplay of morphology and syntax in PLMs, we consider two sets of tasks:
\noindent\textbf{a. Syntactic tasks:} Part-of-speech (POS) tagging, Named Entity Recognition (NER), Dependency Parsing
\noindent\textbf{b. Semantic tasks:} Natural Language Inference(NLI), Paraphrasing\\
%This choice of tasks enables us to understand the function of morphology and word order in encoding syntactic and semantic role information across various languages. Moreover, text classification tasks such as natural language inference and paraphrasing operate at the sentence level. On the other hand, part-of-speech tagging and named entity recognition are sequence labeling tasks that assess the model's competence at a more granular word level. 
Factors considered in task and language selection include (1) availability of monolingual BERT-base model on HuggingFace Hub \citep{huggingface}, (2) availability of sufficient monolingual training data across different tasks, and (3) typological diversity. We aim to cover as many languages and language families as possible. Overall, we cover 22 languages distributed across 9 language families and one language isolate. We present an outline of the languages in Appendix \ref{sec:lang_details} due to space constraints.

\subsection{Datasets}
%In our experiments, we evaluate our models on a variety of tasks. We utilize GLUE \cite{wang2018glue} benchmark which comprises of nine language understanding tasks encompassing a wide spectrum of domains. The benchmark includes sentence acceptability judgment on CoLA~\cite{dataset_cola}, sentiment analysis on SST-2~\cite{dataset_sst2}, paraphrasing/sentence similarity on MRPC~\cite{dataset_mrpc}, STS-B~\cite{dataset_stsb}, QQP~\cite{dataset_qqp}, and  natural language inference on MNLI \cite{dataset_mnli}, QNLI \cite{dataset_qnli}, and RTE \cite{dataset_rte}) datasets. 

%We incorporate multilingual datasets that span various domains and encompass multiple languages. 
Our study includes tasks from the XTREME benchmark \cite{dataset_xtreme}, covering natural language inference (XNLI) \cite{dataset_xnli}, paraphrasing (PAWS-X) \cite{dataset_pawsx}, and structure prediction tasks such as POS tagging and NER. We use the data from the Universal Dependencies v2.12 \citep{dataset_udv2.12} for the task of dependency parsing. The treebanks used for different languages are listed in Table \ref{tab:treebank} in Appendix. %By incorporating these tasks, we aim to enhance the comprehensiveness and breadth of our research.

\subsection{Model Selection} 
In our research, we use monolingual pre-trained language models to prevent cross-lingual transfer from influencing our results. Given the availability of monolingual BERT models in various languages, we select BERT as the example PLM for our study. We consider BERT-base model for all languages to ensure that variations in model size and architecture do not influence the results. We consider fine-tuned BERT-base models with PE and without PE as the baseline and perturbed models, respectively.

\subsection{Evalution Metrics}
The metric used for different tasks is outlined in Table \ref{tab:corr_results}. For a given task, let m and n denote the metric scores for the baseline and perturbed models, respectively. We use the relative decrease in performance, calculated as (m-n)/n, as a quantitative measure of the importance of PE on the language. A higher value indicates a greater utilization of PE in effectively modeling the language. %as an indicator of the importance of positional encoding for the language. %. 
% For classification tasks, we used metrics like
% Matthews Correlation Coefficient for CoLA, Pearson Correlation Coefficient for STS-B and Accuracy for other tasks. For generative tasks, we employed ROUGE (Lin, 2004). We report ROUGE-1,
% ROUGE-2, and ROUGE-L F-scores.
% Let mc and mp denote the values of a metric
% m of the model on the clean and perturbed test
% sets, respectively. Then, we define robustness as
% robustness = 1 −mc−mpmc. Typically, robustness
% score of a model ranges between 0 (not robust)
% and 1 (very robust). Score greater than 1 suggests
% that the model’s performance improves with the
% perturbation applied.

%\subsection{Models}
%In this section, we describe the baseline and perturbed models used in our experiments.

\subsection{Training and evaluation setup}
For text classification tasks, we follow the generic pipeline. For dependency parsing, we implemented a biaffine parser by applying a biaffine attention layer directly on the output of BERT as described in \citet{glavas-vulic-2021-supervised}. As suggested in the XTREME benchmark, we have performed hyper-parameter tuning on English validation data. However, since our goal is not to achieve the best absolute performance, we avoided conducting extensive hyperparameter tuning. More details are present in the Appendix \ref{sec:hp}. Results are reported across 3 random trials of each experiment. %Also, since absolute performance is not the primary objective of this work, we did not perform extensive language-specific hyperparameter tuning. One could likely obtain better scores than what we report in Table 1 with careful language-specific model selection.

% \begin{figure}[!t]
%     \centering
%     \begin{subfigure}[!t]{\linewidth}
%     \centering
%       \includegraphics[width=\textwidth]{acl-style-files-master/Figures/ner_cropped.pdf}
%     \caption{NER}
%     \label{fig:pe_ner}
%     \end{subfigure}
%     \hfill
%     \begin{subfigure}[!t]{\linewidth}
%     \centering
%       \includegraphics[width=\textwidth]{acl-style-files-master/Figures/pos_cropped.pdf}
%     \caption{POS}
%     \label{fig:pe_pos}
%     \end{subfigure}
%     \hfill
%     \begin{subfigure}[!t]{\linewidth}
%     \centering
%       \includegraphics[width=\textwidth]{acl-style-files-master/Figures/uas_cropped.pdf}
%     \caption{Dependency Parsing}
%     \label{fig:pe_dp}
%     \end{subfigure}
    
%     \caption{Effect of Removal Positional Encoding}
%     \label{fig:enter-label}
% \end{figure}

\begin{figure}[!t]
\centering
  \includegraphics[width=\linewidth]{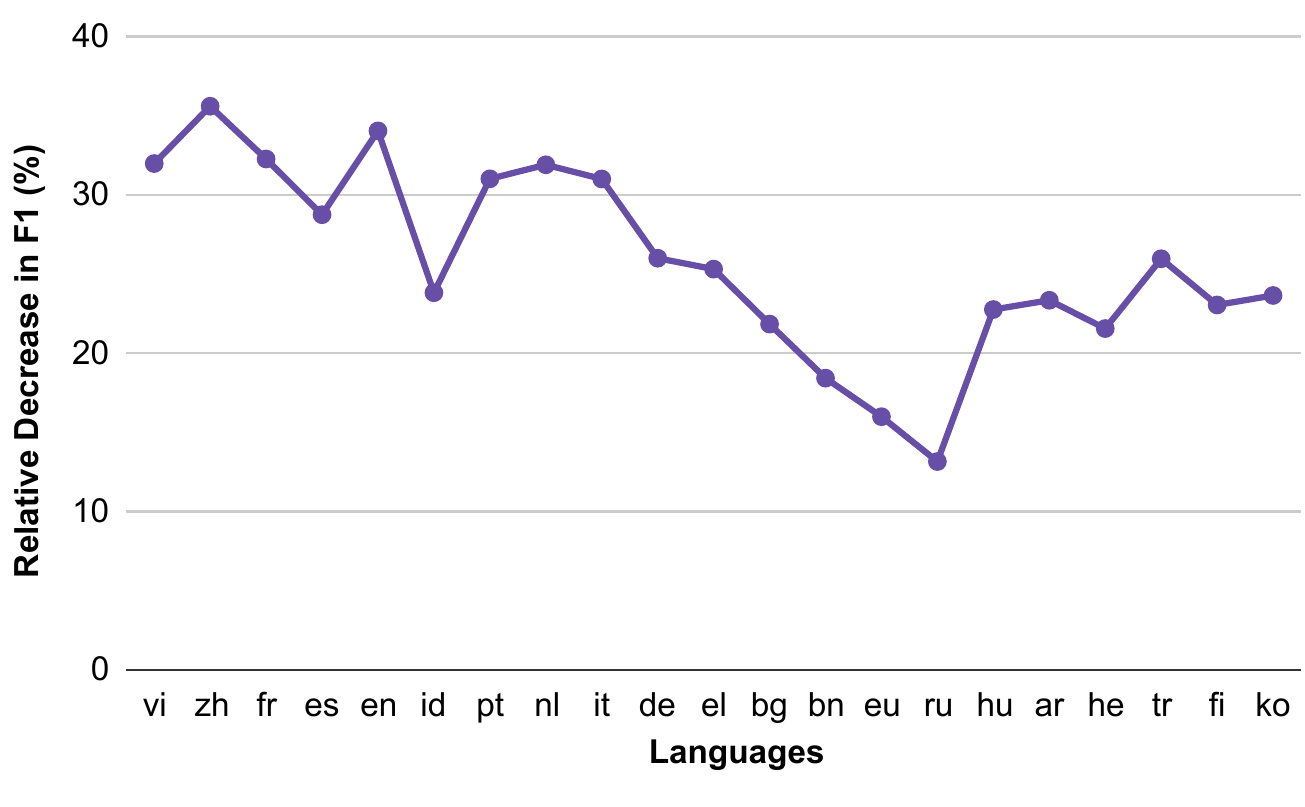}
\caption{Effect of Positional Encoding on NER task. }
\label{fig:pe_ner}
\end{figure}

\begin{figure}[!t]
\centering
  \includegraphics[width=\linewidth]{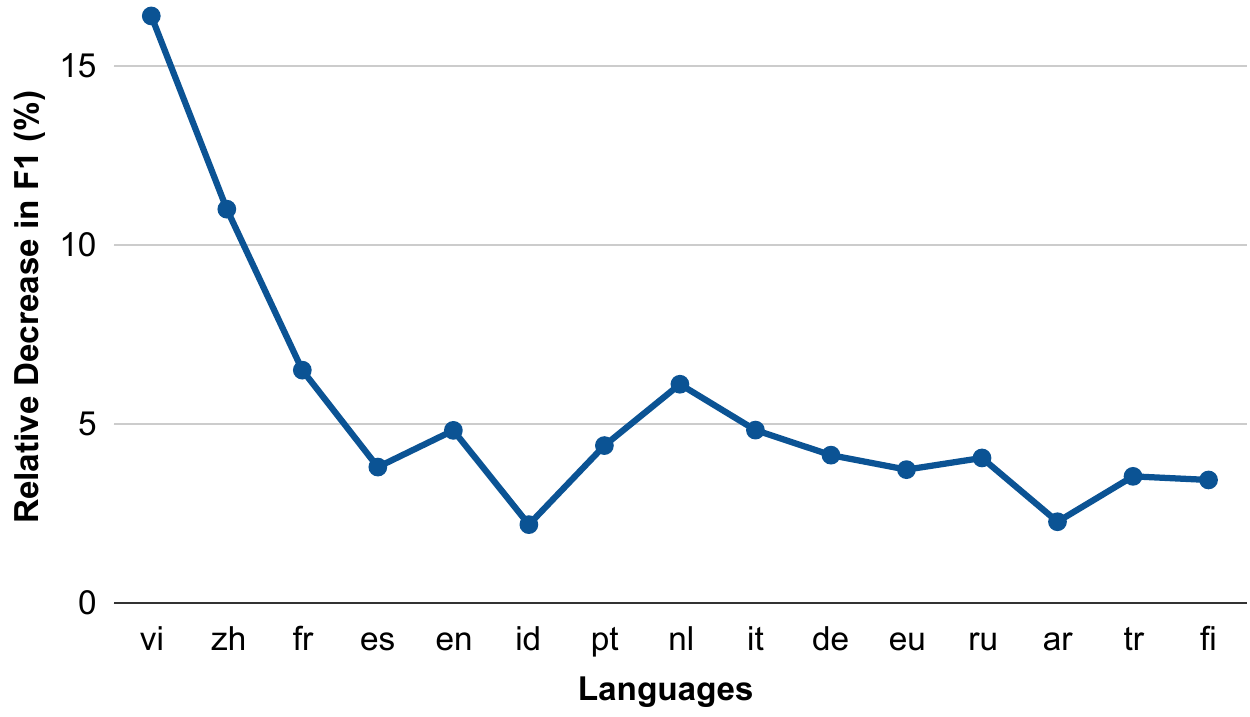}
\caption{Effect of Positional Encoding on POS task.}
\label{fig:pe_pos}
\end{figure}

\begin{figure}[!t]
\centering
  \includegraphics[width=\linewidth]{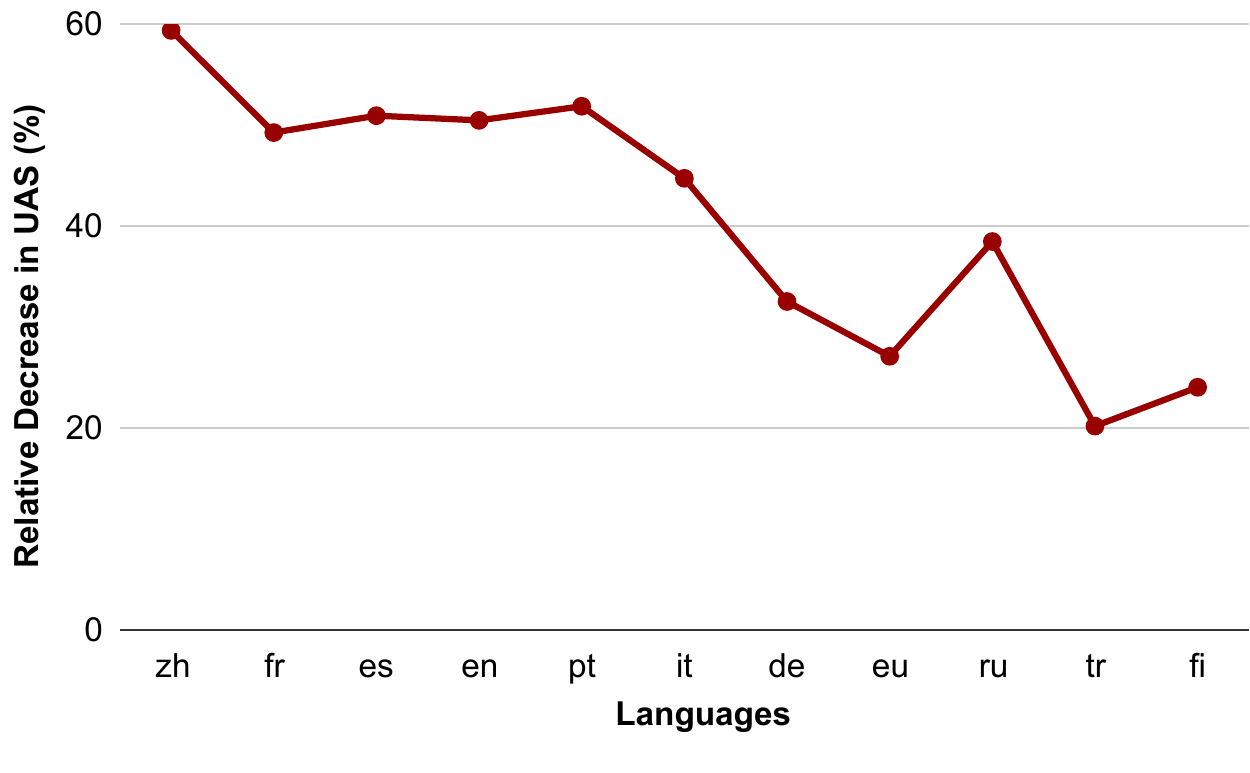}
\caption{Effect of Positional Encoding on Dependency Parsing.}
\label{fig:pe_dp}
\end{figure}

\section{Results}\label{sec:results}
In this section, we present the findings of our experiments on syntactic and semantic tasks. We also conducted preliminary experiments on the GLUE benchmark, as discussed in Appendix \ref{sec:glue_results}. 

%\subsection{Multilingual Analysis and Results}
\subsection{Results on Syntactic Tasks\label{sec:syntactic_results}}
%In this section, we analyze the results for different syntactic tasks.
%\vspace{-4mm}
%\paragraph{Impact of positional encoding:}

Figures \ref{fig:pe_ner}, \ref{fig:pe_pos} demonstrate the effect of removing positional encoding in NER, POS tagging tasks. For dependency parsing, figures \ref{fig:pe_dp} and \ref{fig:pe_pos_las} depict the effects on UAS and LAS scores, respectively.\\
\noindent\textbf{a.} Analytic languages like Chinese and Vietnamese, characterized by minimal or no morphology, exhibit the most significant decrease in performance when PE is removed. Moderately analytic languages like English and French follow.\\
\noindent\textbf{b.} In synthetic languages such as Hungarian, Finnish, and Turkish, known for their rich morphological systems, the function of morphology in encoding grammatical roles surpasses that of word order, resulting in a considerably smaller decrease in performance when PE is eliminated.
%On the other hand, synthetic languages like Finnish, Hungarian, Turkish known for their extensive morphological systems, cluster at the opposite end.
%\item Isolating languages like Chinese and Vietnamese without any inherent morphology, exhibit the most significant decrease in performance when positional encoding is removed. Partially analytic languages with increased inflectional morphology, like English, Italian, and Spanish, follow closely in terms of performance drop. 
%\item At the opposite end of the spectrum, we find fusional languages like German and Russian, as well as agglutinative languages such as Hungarian, Finnish, and Turkish, known for their rich morphological systems. In these languages, the importance of morphology surpasses that of word order, resulting in a considerably smaller decrease in performance when positional encoding is eliminated.

\begin{figure}[!t]
\centering
  \includegraphics[width=\linewidth]{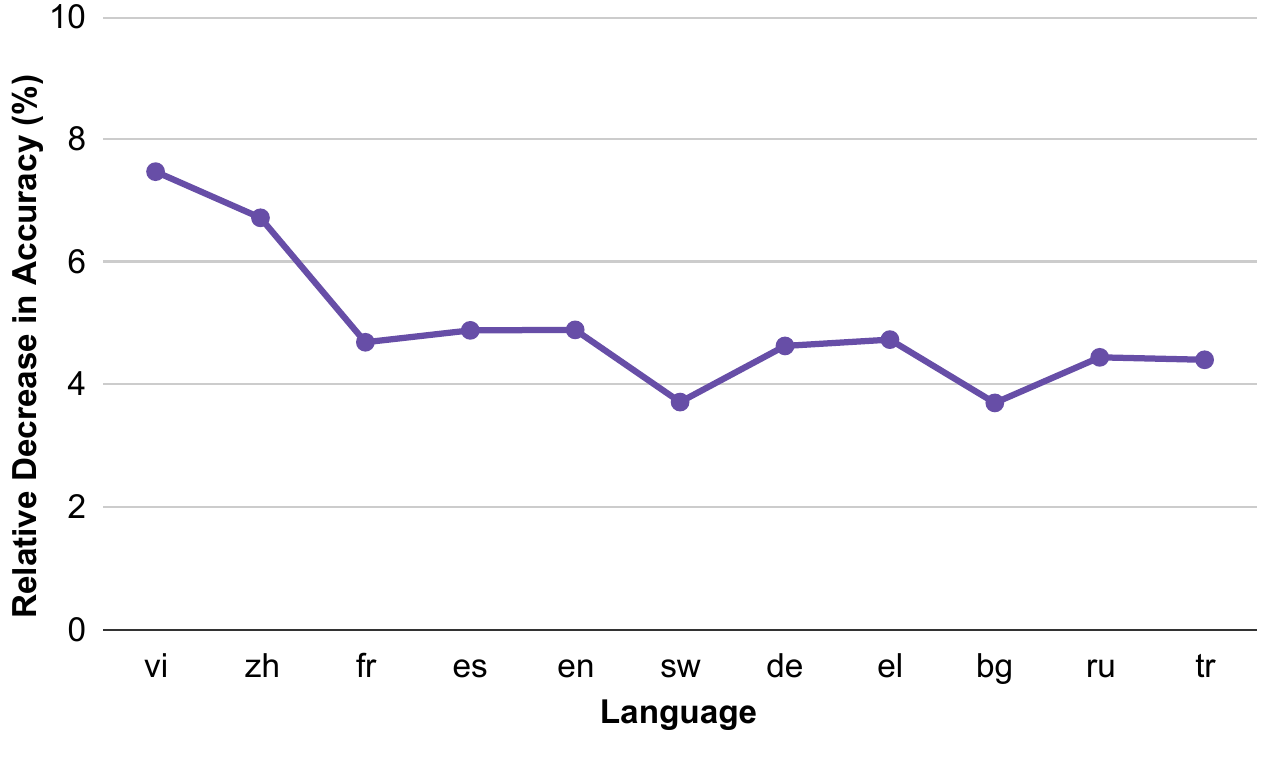}
\caption{Effect of Positional Encoding on XNLI.}
\label{fig:pe_xnli}
\end{figure}

\begin{figure}[!t]
\centering
  \includegraphics[width=\linewidth]{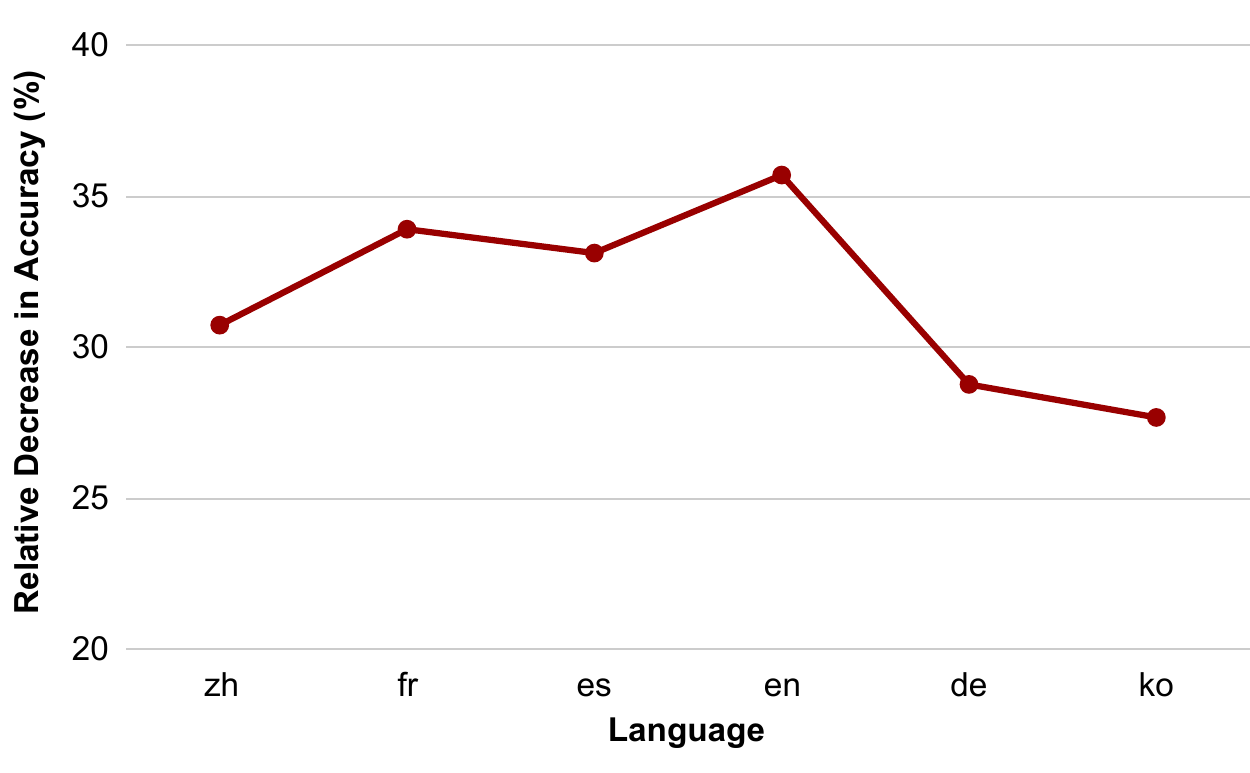}
\caption{Effect of Positional Encoding on PAWS-X.}
\label{fig:pe_paws}
\end{figure}

\noindent In the XTREME benchmark, different methods were employed for data annotation for POS tagging and NER datasets. While the former was human annotated, the latter was created through automatic annotation using weak supervision. Despite these disparities, the findings of the POS tagging and NER experiments are similar.

\subsubsection{Results on Semantic Tasks\label{sec:nli_results}}
Natural language inference and paraphrasing tasks primarily involve understanding the semantic relationships and meaning between sentences or phrases. Syntax, including word order and grammatical rules, can influence the overall coherence and clarity of the expressions, but it is not the primary focus of these tasks.\\ %While syntax can play a role in these tasks to some extent, the main emphasis is on capturing the semantic equivalence or entailment between sentences and generating accurate paraphrases.
The results depicted in Figures \ref{fig:pe_xnli} and \ref{fig:pe_paws} illustrate the impact of nullifying PE in tasks related to natural language inference and paraphrasing. We notice a consistent pattern emerge in the graphs where morphologically poor languages are notably affected by the absence of positional encoding, while the impact is comparatively less for morphologically rich languages. However, in contrast to syntactic tasks, the variability in impact across different languages is less pronounced for semantic tasks.

% \usepackage{booktabs}
\begin{table}[!h]
\center
\begin{tabular}{@{}cc@{}}
\toprule
\textbf{Task (Metric)} & \textbf{Correlation} \\ \midrule
\textbf{NER (F1)} & -0.742 \\
\textbf{POS (F1)} & -0.693 \\
\textbf{Dependency Parsing (UAS)} & -0.882 \\
\textbf{Dependency Parsing (LAS)} & -0.873 \\
\textbf{XNLI (Accuracy)} & -0.773 \\
\textbf{PAWS-X (Accuracy)} & -0.486 \\ \bottomrule
\end{tabular}
\caption{Spearman correlation coefficient between morphological complexity of a language and relative decrease in performance across different tasks}
\label{tab:corr_results}
\end{table}

\noindent In Table \ref{tab:corr_results}, we report the statistical correlation between the morphological complexity of a language and the relative decrease in performance across tasks, as determined by the Spearman Correlation Coefficient. A strong negative correlation is observed, indicating that higher morphological complexity is associated with a lower relative decrease in performance.

% \usepackage{booktabs}
\section{Conclusion}
In this work, we demonstrate the interplay between positional encoding and morphology for morphologically diverse languages. We present the first study regarding the varying impact of positional encoding across languages with varying morphological complexity. Our results reveal that the importance of positional encoding diminishes as the morphological complexity of a language increases. Our study also motivates the need for a deeper understanding of positional encoding, augmenting them to better reflect the different languages under consideration. %By understanding how positional encoding and morphology affect model performance, developers can make more informed decisions about which techniques to use for specific tasks and create more robust models. %Overall, this study contributes to the development of more accurate and robust NLP models that can handle the complexity and diversity of natural languages.
% It is intriguing to observe that morphologically rich languages rely heavily on morphology to convey semantic and syntactic role information, while morphologically poor languages place more emphasis on the position of words in a sentence.

\section*{Ethics Statement}
All the experiments are conducted on openly available datasets and benchmarks with no ethical consideration.

\section*{Limitations}\label{sec:limitations}
Our goal was to incorporate multiple languages to ensure the generalizability of our findings. However, the process of training language models from scratch (without positional encoding) for a large number of languages requires significant computational power and financial resources. We agree that pre-training without positional encoding would be a more holistic approach. However, due to limited computational resources, our primary focus was directed towards fine-tuning. However, we firmly believe that removing positional encoding in the pre-training phase would yield more pronounced results.
%\bibliography{acl_latex}

\appendix
\section{The Relationship between Morphology and Word Order}\label{sec:morph_word_order}
In this section, we investigate the relationship between morphology and word order as outlined in different linguistic theories.\\ \textbf{Linguistic theories :} Morphological case markings serve a similar function as word order \citep{sapir1921language,blake2001case}. Several theories suggest that the presence of morphological case is necessary for free word order in a language \citep{comrie1989language,haspelmath1999optimality}. Either morphological case or structural position facilitates the unambiguous determination of grammatical role of the constituents of a sentence. The existence of morphological case reduces the need for fixed structural position in determination of the grammatical function of a word or phrase, allowing for variable word order.sla However, if morphological case is absent, fixed placement of words (and phrases) is necessary, exhibiting a fixed or rigid word order.
In this work, we align our empirical study in accordance with the above theories that hints at the existence of a correlation between morphology and word order. Specifically, morphologically rich languages which tend to exhibit higher word-order flexibility as compared to morphologically poor languages\footnote{The theory regarding morphology and word order is a linguistically complex topic. Concurrent theories in the literature propose that there is no correlation between morphological complexity and word order. \citet{muller2002free} demonstrates phenomena like scrambling and topic shift, where the change in word order does not necessarily require a high level of morphological complexity. However, this is beyond the scope of our study.}.

\subsection{Spectrum of Morphological Complexity}
Through the lens of morphological typology \cite{haspelmath2013understanding}, we can categorize and cluster languages by studying their inherent morphological structures. At one extreme, we find languages such as Chinese and Vietnamese, which fall into the category of analytic languages and are morphologically impoverished. In these languages, it is essential for words to maintain fixed positions in order to accurately convey grammatical relationships, resulting in a strict and invariant word order. On the other extreme, we find synthetic languages such as Sanskrit and Finnish, known for their rich morphology, where it's possible to rearrange the word order within a sentence without changing its meaning, as illustrated in Figure \ref{fig:overview}. However, most languages fall between these two extremes. Synthetic languages can be categorized into two main types. Agglutinative languages like Hungarian and Turkish tend to stick together multiple morphemes while fusional languages fuse several morphemes to express various grammatical features. %Fusional languages exhibit overloading of grammatical features where a single morpheme can encode multiple grammatical features simultaneously. 
%On the other hand, swapping the order of arguments in languages such as English and French, whose case-marking is limited to relics in the pronominal system, is not possible without swapping their grammatical roles. Latin and Chinook are at one extreme. Such languages are Chinese, Siamese, and Annamite, in which each and every word, if it is to function properly, falls into its assigned place, are at the other extreme. But the majority of languages fall between these two extremes.
%\subsection{Quantifying Morphological Complexity}\label{sec:morphological_complexity}
%We perform standard tokenization using nltk library for space-delimited languages. No need to mention about nltk.

\section{Details of Languages}\label{sec:lang_details}
We provide an overview of the languages included in our study in Table \ref{tab:languages}. Additionally, Table \ref{tab:treebank} presents the details of the treebanks used in the dependency parsing experiments. 

\section{TTR-based Morphological Complexity}\label{sec:ttr_morph}
The TTR-based morphological complexity of different languages are listed in Table \ref{tab:ttr}. For space-delimited languages, we use the tokenizer from NLTK library for word segmentation.

% \usepackage{booktabs}
\begin{table*}[]
\begin{tabular}{@{}ccc@{}}
\toprule
\textbf{Language (ISO code)} & \textbf{Language family} & \textbf{Hugging Face Model id} \\ \midrule
Arabic (ar) & Afro-Asiatic & aubmindlab/bert-base-arabertv02 \\
Basque (eu) & Basque & orai-nlp/ElhBERTeu \\
Bengali (bn) & Indo-European: Indo-Aryan & sagorsarker/bangla-bert-base \\
Bulgarian (bg) & Indo-European: Slavic & usmiva/bert-web-bg \\
Chinese (zh) & Sino-Tibetan & bert-base-chinese \\
Dutch (nl) & Indo-European: Germanic & GroNLP/bert-base-dutch-cased \\
English (en) & Indo-European: Germanic & bert-base-cased \\
Finnish (fi) & Uralic & TurkuNLP/bert-base-finnish-cased-v1 \\
French (fr) & Indo-European: Romance & dbmdz/bert-base-french-europeana-cased \\
German (de) & Indo-European: Germanic & dbmdz/bert-base-german-cased \\
Greek (el) & Indo-European: Greek & nlpaueb/bert-base-greek-uncased-v1 \\
Hebrew (he) & Afro-Asiatic & onlplab/alephbert-base \\
Hungarian (hu) & Uralic & SZTAKI-HLT/hubert-base-cc \\
Indonesian (id) & Austronesian & indolem/indobert-base-uncased \\
Italian (it) & Indo-European: Romance & dbmdz/bert-base-italian-cased \\
Korean (ko) & Koreanic & kykim/bert-kor-base \\
Portuguese (pt) & Indo-European: Romance & neuralmind/bert-base-portuguese-cased \\
Russian (ru) & Indo-European: Slavic & DeepPavlov/rubert-base-cased \\
Spanish (es) & Indo-European: Romance & dccuchile/bert-base-spanish-wwm-cased \\
Swahili (sw) & Niger-Congo & flax-community/bert-base-uncased-swahili \\
Turkish (tr) & Turkic & dbmdz/bert-base-turkish-cased \\
Vietnamese (vi) & Austro-Asiatic & trituenhantaoio/bert-base-vietnamese-uncased \\ \bottomrule
\end{tabular}
\caption{Overview of different languages}
\label{tab:languages}
\end{table*}

% \usepackage{booktabs}
\begin{table*}[]
\centering
\begin{tabular}{@{}ccc@{}}
\toprule
\textbf{Language (ISO code)} & \textbf{FLORES-200 code} & \textbf{TTR} \\ \midrule
Arabic (ar) & arb\_Arab & 0.359 \\
Basque (eu) & eus\_Latn & 0.324 \\
Bengali (bn) & ben\_Beng & 0.292 \\
Bulgarian (bg) & bul\_Cyrl & 0.268 \\
Chinese (zh) & zho\_Hans & 0.17 \\
Dutch (nl) & nld\_Latn & 0.207 \\
English (en) & eng\_Latn & 0.194 \\
Finnish (fi) & fin\_Latn & 0.428 \\
French (fr) & fra\_Latn & 0.191 \\
German (de) & deu\_Latn & 0.244 \\
Greek (el) & ell\_Grek & 0.253 \\
Hebrew (he) & heb\_Hebr & 0.364 \\
Hungarian (hu) & hun\_Latn & 0.345 \\
Indonesian (id) & ind\_Latn & 0.195 \\
Italian (it) & ita\_Latn & 0.217 \\
Korean (ko) & kor\_Hang & 0.465 \\
Portuguese (pt) & por\_Latn & 0.205 \\
Russian (ru) & rus\_Cyrl & 0.334 \\
Spanish (es) & spa\_Latn & 0.192 \\
Swahili (sw) & swh\_Latn & 0.212 \\
Turkish (tr) & tur\_Latn & 0.376 \\
Vietnamese (vi) & vie\_Latn & 0.077 \\ \bottomrule
\end{tabular}
\caption{TTR-based morphological complexity of different languages}
\label{tab:ttr}
\end{table*}

\begin{table*}[]
\centering
\begin{tabular}{@{}ll@{}}
\toprule
\multicolumn{1}{c}{\textbf{Language}} & \multicolumn{1}{c}{\textbf{Treebank}} \\
\midrule
Chinese (zh) & UD\_Chinese-GSD \\
Portuguese (pt) & UD\_Portuguese-Bosque \\
Spanish (es) & UD\_Spanish-GSD \\
English (en) & UD\_English-GUM \\
French (fr) & UD\_French-GSD \\
Italian (it) & UD\_Italian-ISDT \\
Russian (ru) & UD\_Russian-Taiga \\
German (de) & UD\_German-GSD \\
Basque (eu) & UD\_Basque-BDT \\
Finnish (fi) & UD\_Finnish-FTB \\
Turkish (tr) & UD\_Turkish-Penn \\ 
\bottomrule
\end{tabular}
\caption{Details of treebanks of different languages}
\label{tab:treebank}
\end{table*}

\section{Additional Results}\label{sec:dp_las}
\begin{figure}[!h]
\centering
  \includegraphics[width=\linewidth]{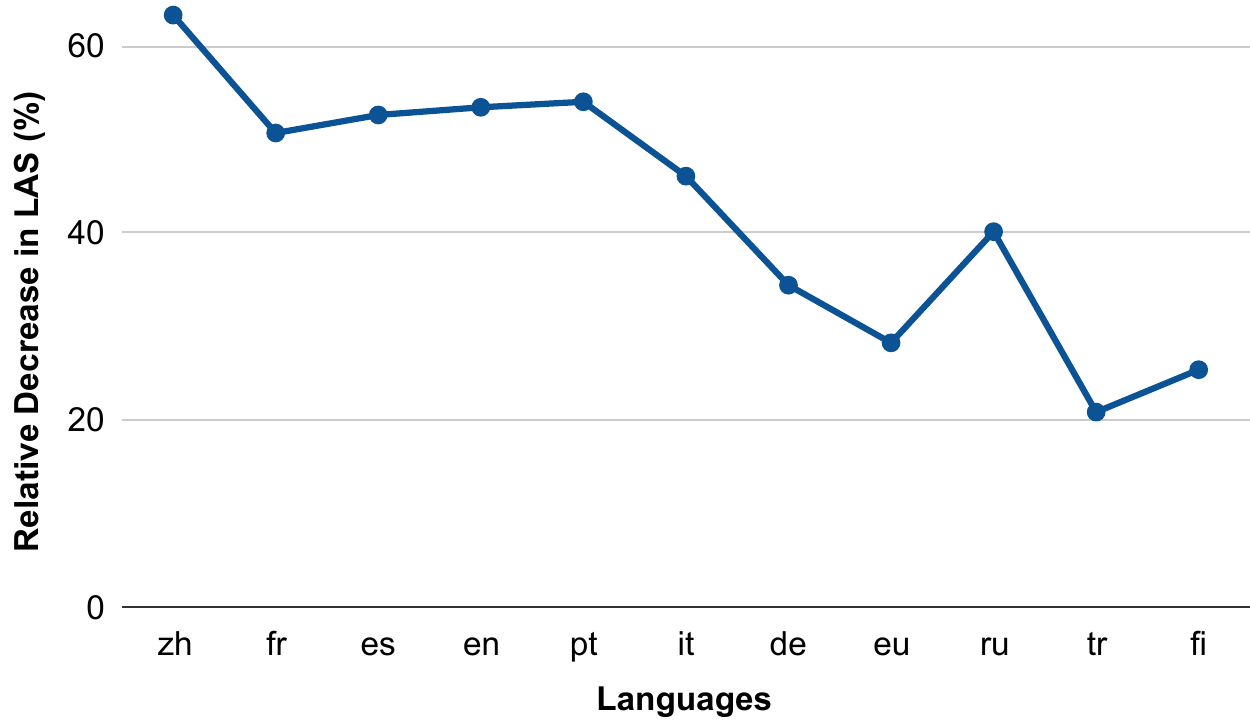}
\caption{Effect of Positional Encoding on Dependency Parsing (LAS).}
\label{fig:pe_pos_las}
\end{figure}

The effect of removing positional encoding in dependency parsing is examined by analyzing the relative decrease in UAS (Figure \ref{fig:pe_dp}) and LAS scores (Figure \ref{fig:pe_pos_las}).

\section{Results on GLUE tasks}\label{sec:glue_results}
\begin{table*}[]
\centering
\resizebox{\textwidth}{!}{%
\begin{tabular}{cccccccc}
\toprule
\multirow{3}{*}{\textbf{Task}} & \multirow{3}{*}{\textbf{Dataset}} & \multicolumn{4}{c}{\textbf{With Positional Encoding}} & \multirow{3}{*}{\textbf{\begin{tabular}[c]{@{}c@{}}Without \\ Positional \\ Encoding\end{tabular}}} & \multirow{3}{*}{\textbf{\begin{tabular}[c]{@{}c@{}}Relative \\ decrease\end{tabular}}} \\ 
\cmidrule{3-6} 
 &  & \textbf{Unigram} & \textbf{Bigram} & \textbf{Trigram} & \textbf{Without}&  &  \\
 &  & \textbf{Permutation} & \textbf{Permutation} & \textbf{Permutation} & \textbf{Permutation}&  &  \\ \midrule
\textbf{\begin{tabular}[c]{@{}c@{}}Sentence\\ Acceptability\end{tabular}} & \textbf{CoLA} & \multicolumn{1}{c}{4.4} & \multicolumn{1}{c}{12.3} & \multicolumn{1}{c}{16.8} & 59.1 & 23.8 & 59.7 \\ \midrule
\textbf{\begin{tabular}[c]{@{}c@{}}Sentiment \\ Analysis\end{tabular}} & \textbf{SST-2} & \multicolumn{1}{c}{81.6} & \multicolumn{1}{c}{86.0} & \multicolumn{1}{c}{85.1} & 91.8 & 86.5 & 5.8 \\ \midrule
\multirow{3}{*}{\textbf{\begin{tabular}[c]{@{}c@{}}Paraphrasing / \\ Sentence \\ Similarity\end{tabular}}} & \textbf{MPRC} & \multicolumn{1}{c}{83.5} & \multicolumn{1}{c}{84.2} & \multicolumn{1}{c}{85.4} & 89.8 & 84.6 & 5.8 \\ 
 & \textbf{QQP} & \multicolumn{1}{c}{75.6} & \multicolumn{1}{c}{79.1} & \multicolumn{1}{c}{80.8} & 87.1 & 85.8 & 1.5 \\ 
 & \textbf{STS-B} & \multicolumn{1}{c}{85.2} & \multicolumn{1}{c}{87.1} & \multicolumn{1}{c}{86.6} & 89.0 & 86.6 & 2.7 \\ \midrule
\multirow{4}{*}{\textbf{\begin{tabular}[c]{@{}c@{}}Natural \\ Language \\ Inference \end{tabular}}} & \textbf{MNLI} & \multicolumn{1}{c}{68.3} & \multicolumn{1}{c}{74.8} & \multicolumn{1}{c}{76.5} & 83.6 & 79.7 & 4.7 \\ 
 & \textbf{MNLI-MM} & \multicolumn{1}{c}{68.7} & \multicolumn{1}{c}{74.4} & \multicolumn{1}{c}{76.6} & 84.0 & 79.8 & 5.0 \\ 
 & \textbf{QNLI} & \multicolumn{1}{c}{81.3} & \multicolumn{1}{c}{85.0} & \multicolumn{1}{c}{86.5} & 91.0 & 87.2 & 4.2 \\ 
 & \textbf{RTE} & \multicolumn{1}{c}{58.1} & \multicolumn{1}{c}{61.5} & \multicolumn{1}{c}{61.8} & 64.5 & 62.8 & 2.6 \\
 \bottomrule
\end{tabular}%
}
\caption{GLUE Results for English language: The evaluation metrics used for reporting the performance of QQP and MRPC tasks are F1 scores, while for the STS-B task, Spearman correlations are used, and accuracy scores are employed for the remaining tasks. The average and standard deviation are reported across 3 trails of the evaluation on the validation set. The relative decrease quantifies the decline in performance when positional encoding is excluded compared to when positional encoding was present. Additionally, we conducted experiments in which we removed positional encoding and perturbed the input to the model. Since the removal of positional encoding results in a bag of words model, we observed no noticeable change upon further distortion.}
\label{tab:glue_results}
\end{table*}

\subsubsection{Impact of positional encoding}
Removing positional encoding leads to a varied decrease in performance across different tasks, as evident in Table \ref{tab:glue_results}.\\% of the GLUE benchmark.
\textbf{Sentence/Grammatical Acceptability Tasks:}  Positional encoding helps the model understand the hierarchical structure and dependencies between words, which is essential for determining the grammaticality of a sentence. As a result, in case of CoLA task, when positional encoding is removed, the model struggles to identify grammatically acceptable sentences, leading to a notable decline of 59.7\% in performance.\\
\textbf{Paraphrasing and sentence similarity Tasks:} Models can effectively capture the similarity or relatedness between sentences by focusing on common signals present across sentences. These tasks primarily require understanding the underlying semantic meaning and contextual similarities between sentences rather than the syntactic structure.As a result, when positional encoding is removed, the relative decrease in performance is considerably smaller (5.8\% for MRPC, 2.7\% for STS-B, and 1.5\% for QQP). This indicates that while positional encoding does provide some benefit in capturing the positional information within sentences, it is not very crucial for these tasks.\\
\textbf{Natural language inference Tasks:} The removal of positional encoding leads to a relative decrease of 4.7\% for MNLI, 5\% for the mismatched version of MNLI,  4.2\% for QNLI, and 2.6\% for RTE. The decrease in performance is still relatively modest for these tasks. This highlights that positional encoding does not play a significant role in understanding the logical inference and entailment relationships between sentence pairs \cite{wang2020position}.\\
Even in the absence of positional encoding, the bag of words model gives  considerably good performance for paraphrasing and natural language inference tasks.The results on GLUE benchmark serve as a driving force behind our investigation, where we aim to further test our hypothesis across morphologically diverse languages.

\subsection{Impact of permutation:}
In addition to studying the effect of positional encoding, we also conducted experiments to examine the impact of permutation on various GLUE tasks.
\begin{itemize}
[itemsep=2pt,parsep=2pt,partopsep=2pt,leftmargin=*,topsep=2pt]
    \item Unigram permutation causes the most significant performance drop. However, as we increase the ngram order, which involves shuffling chunk of words instead of individual words, the decrease in performance is significantly less. This indicates that shuffling at higher ngram levels add less distortion and preserve the integrity of word order to a greater extent.
    \item The results also imply that lower order ngrams capture vocabulary match and is completely ignorant of word order whereas higher order ngrams capture word order and other dependencies present in a sentence.
\end{itemize}

\section{Hyper-parameter details:}\label{sec:hp}
The hyper-parameter details used at the time of fine-tuning are outlined in Table \ref{tab:hp}. For dependency parsing, we have followed the hyper-parameter settings mentioned in \citet{glavas-vulic-2021-supervised}.
% \usepackage{booktabs}
\begin{table*}[]
\centering
\begin{tabular}{@{}cccc@{}}
\toprule
\textbf{Task} & \textbf{learning rate} & \textbf{batch size} & \textbf{number of epochs} \\ \midrule
NER & 2.00E-05 & 16 & 3 \\
POS & 3.00E-05 & 8 & 3 \\
XNLI & 3.00E-05 & 32 & 3 \\
PAWS-X & 3.00E-05 & 32 & 3 \\ \bottomrule
\end{tabular}
\caption{Hyper-parameter details}
\label{tab:hp}
\end{table*}

\end{document}